\pdfoutput=1

\documentclass[11pt]{article}

\usepackage{parskip}

\usepackage[]{EMNLP2023}

\usepackage{times}
\usepackage{latexsym}

\usepackage[T1]{fontenc}

\usepackage[utf8]{inputenc}

\usepackage{microtype}

\usepackage{inconsolata}

\usepackage{graphicx}
\usepackage{listings}
\usepackage{xcolor}
\usepackage{outlines}
\usepackage{multirow}

\title{Utilizing Weak Supervision to Generate Indonesian Conservation Datasets}

\author{Mega Fransiska \\ 
    \texttt{mega.fransiska@datasaur.ai} \\\And 
    Diah Pitaloka \\ 
    \texttt{diah.pitaloka@datasaur.ai}  \\\AND 
    Saripudin \\ 
    \texttt{saripudin@datasaur.ai} \\\And 
    Satrio Putra \\ 
    \texttt{satrio.putra@datasaur.ai} \\\And 
    Lintang Sutawika* \\ 
    \texttt{lintang@eleuther.ai}\thanks{*Work completed in Datasaur.ai} \AND
    Datasaur.ai \And 
    *Eleuther.ai \\
}

\begin{document}
\maketitle

\begin{abstract}
\begin{sloppypar}
    Weak supervision has emerged as a promising approach for rapid and large-scale dataset creation in response to the increasing demand for accelerated NLP development. By leveraging labeling functions, weak supervision allows practitioners to generate datasets quickly by creating learned label models that produce soft-labeled datasets. This paper aims to show how such an approach can be utilized to build an Indonesian NLP dataset from conservation news text. We construct two types of datasets: multi-label classification and sentiment classification. We then provide baseline experiments using various pretrained language models. These baseline results demonstrate test performances of 59.79\% accuracy and 55.72\% F1-score for sentiment classification, 66.87\% F1-score-macro, 71.5\% F1-score-micro, and 83.67\% ROC-AUC for multi-label classification. Additionally, we release the datasets and labeling functions used in this work for further research and exploration. \newline
\end{sloppypar}
\end{abstract}

\section{Introduction}

Labeled datasets play a crucial role in Natural Language Processing (NLP) tasks. However, generating large-scale labeled datasets with high quality remains a significant challenge. In addressing this challenge, weak supervision has emerged as a promising approach \cite{Ratner2016DatProgramming, Zhang:2022} to create labeled datasets by leveraging weak classifiers and aggregating their outputs into noisy labels that approximate the unobserved ground truth. Empirical studies \cite{Ratner:2019, Ratner:2020, Ren2020MultiWS, lan-etal-2020-learning, Lison2021skweakWS} have demonstrated the competitiveness of this approach compared to manual data collection processes. Additionally, benchmarks \cite{zhang2021wrench} have been established to further evaluate new weak-supervision approaches. \\
The idea of weak supervision methods to create datasets is something that can be useful for under-resourced languages such as Indonesian. The current approach to manually label every sample is a painful process that realistically results in relatively small amounts of data when budgets are limited. Another approach could be crowdsourcing  \cite{Cahyawijaya2022NusaCrowdOS}, but this approach rests upon many contributors whose quality or theme can be vastly different. To add more resources from downstream tasks in the Indonesian language, we seek to build datasets at larger sizes by using weak supervision. \\
In this paper, we utilize Datasaur's weak supervision framework, Data programming, to facilitate the creation of multiple diverse Indonesian downstream datasets with conservation as the overarching theme. Our focus is leveraging the Mongabay conservation news dataset into two NLP tasks: multi-label classification and sentiment classification. We construct a hashtag classification dataset for multi-label classification, considering the importance of hashtag classification in organizing and enhancing searchability and granularity in editorial realm\footnote{\url{https://whatsnewinpublishing.com/the-importance-of-tags-in-online-news-media/}}. Then, we also use utilize hashtag classification dataset to build sentiment classification dataset by categorizing groups of hashtags into related sentiment categories (positive, neutral, negative). For building both datasets (hashtag classification and sentiment classification), we employ a range of simple labeling functions \cite{Ratner:2020} through Datasaur's Data Programming. \\
Our methodology encompasses dataset construction, weak-labeled dataset learnability experiments through various BERT pre-trained models, and performance analysis of labeling functions. We further elaborate our approach, outlining dataset and labeling functions collections, along with the associated benefits and constraints. \\

\graphicspath{{images/}}
\begin{figure*}[htbp]
    \centering
    \includegraphics[width=15cm]{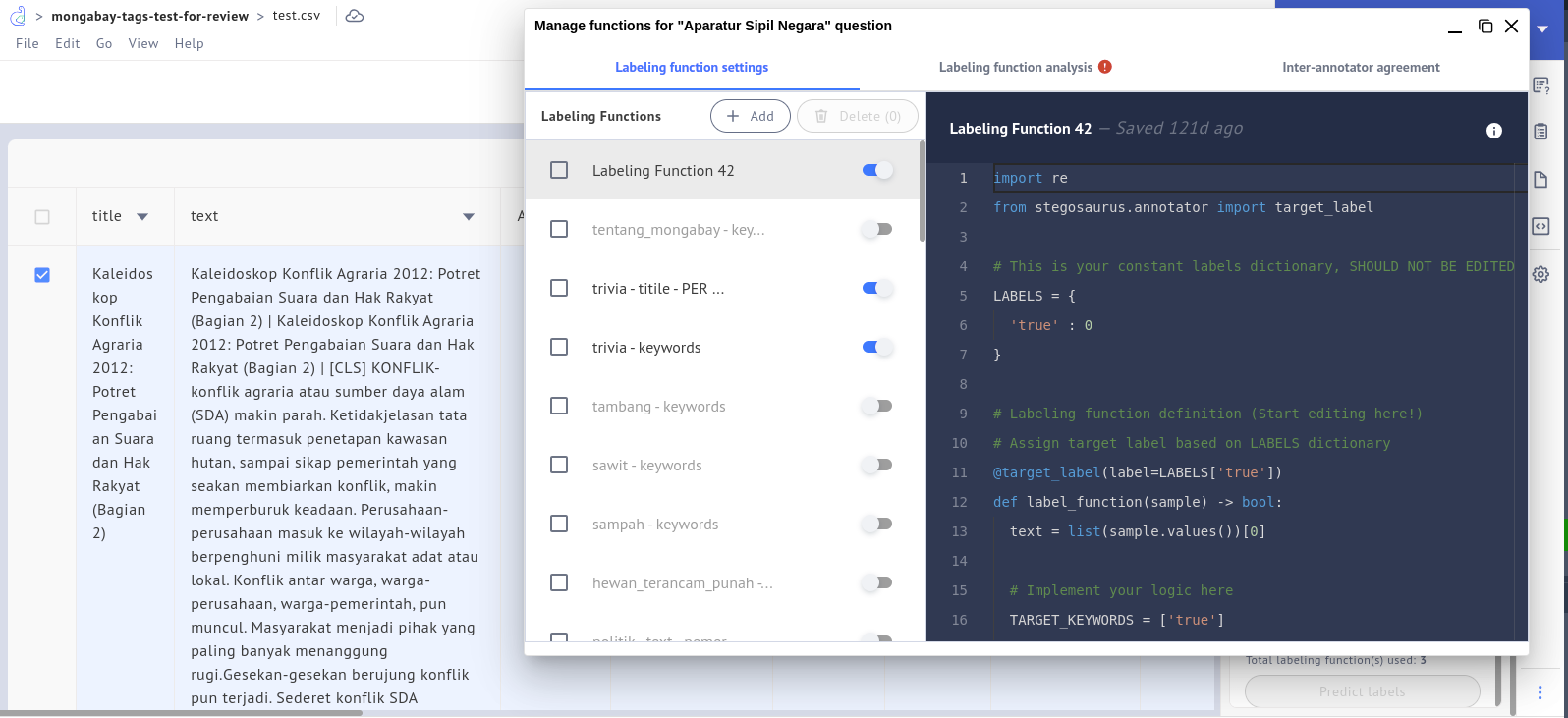}
    \caption{Integrated Labeling function editor in Datasaur workspace}
    \label{fig:data_programming}
\end{figure*}

\section{Related Work}
\textbf{Weak Supervision}. Machine learning requires a large amount of labeled data, which can be costly and difficult to scale. To reduce the cost of building a dataset, weak supervision uses a heuristic approach to generate a training set with noisy labels from multiple sources \cite{Ratner2016DatProgramming, Ratner:2020, Alexander_2022}. The result is a dataset with probabilities as labels or what can be called soft labels. The idea behind this approach is to encode knowledge from experts into labeling functions. These labeling functions serve as a weak classifier where they individually cannot yield a good prediction but when used in tandem with many other label functions can be an effective approximation to the unobserved ground truth. The process is then concluded with a generative model that aims to model the labeling functions by taking into account the agreement and disagreement between labeling functions \cite{Ratner2016DatProgramming, Ratner:2020, Alexander_2022} (later on, this generative model will be called \textbf{label model}). The output of a label model is a noisy signal that estimates the true labels which can be used to predict the soft-labels of a sample. The results have shown considerable gains and are highly applicable in real-world industries for reducing the cost of hand-labeled data \cite{Ratner2016DatProgramming, Ratner:2020}.

These related works used \cite{Ratner2016DatProgramming} version of weak supervision to build training sets. In \citet{Ratner2016DatProgramming}, experiments were conducted using the 2014 TAC-KBP Slot Filling dataset. The results achieved 6 F1 points over a state-of-the-art LSTM. In another study, weak supervision was used to build the ORCAS dataset for user intent classification tasks \cite{Alexander_2022}. The ORCAS dataset consists of query ID, query, document ID, and clicked URL. To label the dataset, a 2-million sample of data was used. The researchers conducted experiments with machine learning models and evaluated the results. The findings indicated competitive results and high efficiency in real-world problems, where labeling functions can be easily executed for every query issued. Finally, a user study showed that using the pipeline of a weak supervision framework can increase predictive performance even faster than seven hours of hand labeling \cite{Ratner:2020}.

\textbf{Sentiment Classification}. Sentiment classification is a classification task to extract the polarity or sentiment expression in text \cite{davidov-etal-2010-enhanced}. Several works used data from social media, such as Twitter, for classification tasks and leveraged the hashtags as additional features alongside the content of the tweet \cite{davidov-etal-2010-enhanced, Devi2019TrendingtagsClassificationP, Diao2023HashtagGuidedLT}. \\
\cite{davidov-etal-2010-enhanced} proposed a supervised sentiment classification framework using Twitter tags and 15 smileys as features. The results showed good performance in labeling data without manual annotation. Another study \cite{Devi2019TrendingtagsClassificationP} explored the hashtag and content tweets to predict which hashtags will become trending in the future. By using machine learning approaches, hashtags as features contribute the better results of the model. Also \cite{Diao2023HashtagGuidedLT} uses hashtags to provide auxiliary signals to get labels for the data. They generated hashtags from input text to produce new input for the model, with meaningful hashtags. The hashtag generator uses an encoder and decoder to predict the hashtags, and the results showed significant improvements in tweet classification tasks.

\textbf{Multi-Label Classification}. In comparison to other text classification research, the field of multi-label classification remains relatively underexplored. Nonetheless, it constitutes a valuable NLP task for extracting metadata from extensive textual datasets, like research papers and articles. For instance, \citet{Li_2021_multilabel} leveraged a KNN-based model to address the challenges posed by multi-label classification in the context of research papers.

\section{Building Dataset Using Weak Supervision}
\label{sec:general_data_programming_work}
Building large datasets using weak supervision has been demonstrated to be effective and of high quality in many studies. For instance, in a recent study by \citet{Tekumalla2022TweetDis}, they curated a silver standard dataset
(samples of raw data sources that have good enough quality to be trained, which were collected and cleaned using weak supervision heuristics) for natural disasters using weak supervision. Similarly, another study by \citet{Painter2022UtilizingWS} utilized weak supervision to create a silver-standard sarcasm-annotated dataset (S3D) containing over 100,000 tweets. This approach holds great promise for expanding the availability of labeled datasets, facilitating the development of more accurate and robust machine learning models.

Our research is focused on curating Indonesian conservation datasets using a weak supervision framework, which has been adapted from Snorkel's works \cite{Ratner2016DatProgramming, Ratner:2020, Alexander_2022}. To make the process as user-friendly as possible, we have developed interactive weak supervision tools, known as Data Programming, in Datasaur\footnote{\url{https://datasaur.ai/}} workspace. Our Data Programming is integrated with a simple code editor in the workspace, which allows users to create labeling functions interactively (figure \ref{fig:data_programming}). We've also provided a Python labeling function template, as detailed in \ref{sec:lf_appendix}. The predictions generated by each labeling function are processed in the background by a label model \cite{Ratner:2019}.

Our Data Programming returns two types of results: probability outputs, which are used as soft labels in the fine-tuning process, and hard-label predictions. These hard-label predictions can be reviewed and revised by human annotators directly in our workspace. 
In this work, we use the probability results in our training set and revise the hard-label prediction for validation and test set, as our \textbf{golden-set}.

\subsection{Dataset Source}
\begin{sloppypar}
    We collect articles from the Indonesian conservation news collection,  Mongabay\footnote{\url{https://www.mongabay.co.id/}}, in 2012 - 2023 period. The raw dataset was sampled from either the first or last 100 articles in each year. These articles were then segmented into multiple chunks, with each chunk containing a maximum of 512 tokens, The format for each data point is as follows:

    \verb|{title} ; {chunked-article}|
\end{sloppypar}

This process resulted in a total of 4896 chunked articles, which were split into 3919, 492, and 485 chunked articles as training set, validation set, and test set respectively.

\subsection{Task}
We categorized the scraped dataset into two primary tasks: multi-label classification and sentiment classification. The multi-label classification task aimed to depict the distribution of hashtags within the dataset. On the other hand, the sentiment classification task was undertaken to gauge the sentiment of the authors, which was still embedded in the articles. For the sentiment classification task, we constructed it based on the hashtag distribution dataset, as previously employed in these works \cite{Devi2019TrendingtagsClassificationP, davidov-etal-2010-enhanced}.

\graphicspath{{images/}}
\begin{figure*}[htbp]
    \centering
    \includegraphics[width=10cm]{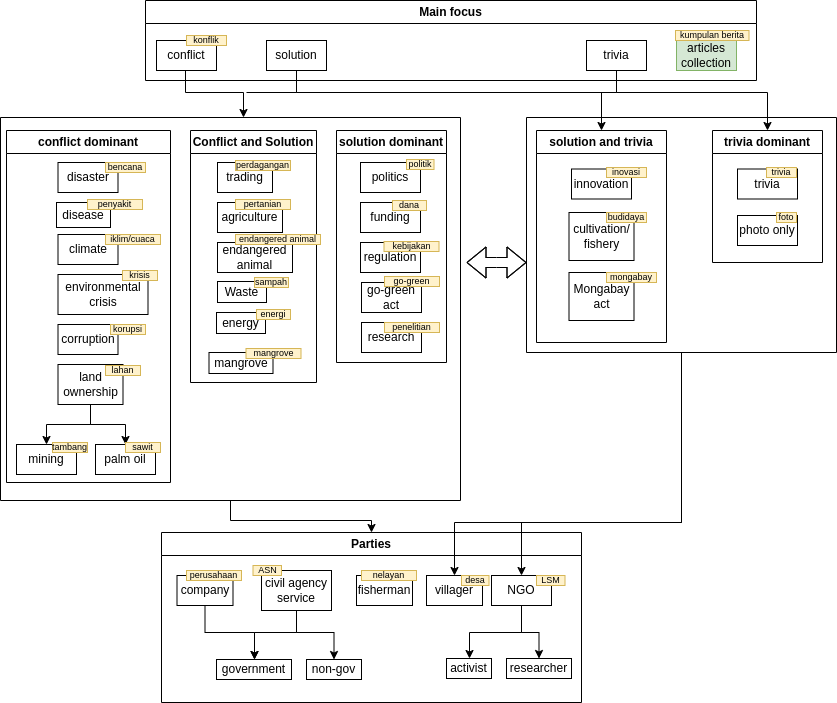}
    \caption{The representation of 31 tags in our dataset; yellow box is our tag label and green box is special class because the article consists of many different articles.}
    \label{fig:tags_structure}
\end{figure*}

We defined 31 classes for the hashtag classification task and 3 classes for the sentiment classification task. The 31 tags were collected by our labelers through an internal analysis of popular environmental and news topics among Indonesian citizens, as shown in Figure \ref{fig:tags_structure}. We acknowledge that this approach heavily depends on the knowledge and personal experience of the labelers in the field of conservation news and environmental topics.
The subsequent section (Section \ref{sec:dataset_construction}) will offer more comprehensive insights into the dataset construction.

\subsection{Dataset Construction}
\label{sec:dataset_construction}
As mentioned in Section \ref{sec:general_data_programming_work}, our data programming generates hard labels and probability labels. In this work, our strategy involves using data programming for labeling the entire dataset, including the training, validation, and test sets. We use the probability outputs as our training set, while our validation and test sets use hard-label outputs that have been reviewed and revised by human annotators, forming our \textbf{golden-set}. \\
When reviewing the hard-label predictions, our labelers follow this simple guideline:
\begin{itemize}
    \item \textbf{Hashtag Classification}: A hashtag is assigned to a chunked article if it is discussed in the text, even if it's mentioned as a side effect.
    \item \textbf{sentiment Classification}: Chunked articles are labeled as follows:
    1) \textbf{negative:} If they mention any conflict or victims. 2) \textbf{neutral:} If there is no discernible sentiment tone, the article is purely descriptive, OR it contains both a conflict and its resolution. 3) \textbf{positive:} If the article discusses trivia topics or initiatives aimed at solving environmental issues.
\end{itemize}
The dataset construction processes for both hashtag classification and sentiment classification involve three key steps: \textbf{labeling function construction}, \textbf{labeling function analysis}, and \textbf{label model and final prediction}.

\subsubsection{Labeling Function Construction}
\label{sec:lf_construction}
The labeling functions used in this study are based on collected keywords from the labeler's perspective. In certain cases, additional rules and logic were added to augment the labeling functions. For the hashtag classification task, labelers gathered relevant keywords for each collected hashtag. In the sentiment classification task, the labeling functions relied on aggregated tags on each article corresponding to positive, neutral, or negative labels.

For the sentiment classification task, we developed two versions of labeling functions: the default version, v0, which was used in the main experiment, and v1 which has more specific logic. The detailed methodology for building labeling functions in the tags classification task and sentiment classification task undertaken in this work is presented in Appendix \ref{sec:lf_appendix}.

\subsubsection{Labeling Function Analysis}
\label{sec:label_function_analysis}

To evaluate the performance of sentiment classification labeling functions, we used coverage, overlaps, and conflict statistics, which have been defined in \citet{Ratner2016DatProgramming}. However, in tags classification, we only utilize coverage score to represent the density of tags in each article, as other metrics such as overlaps and conflict did not adequately reflect the quality of the labeling functions. \\
As discussed in Section \ref{sec:lf_construction}, for the sentiment classification task, we developed v1 labeling functions with more specific logic. This resulted in higher level of inter-independence among the labeling functions, leading to lower coverage and conflict scores (Table \ref{tab:lf_analysis}). Notably, the v1 labeling functions exhibited a significantly smaller percentage of conflict/coverage (2.9\%) compared to v0 (32\%). \\
As highlighted in \citet{Ratner2016DatProgramming}, the statistical performance of the labeling functions directly impacts the quality of the final label prediction and the performance of fine-tuned models. Hence, we conducted experiments and analyzed the influence of the quality of two labeling function versions (v0: prioritizing coverage score; v1: prioritizing less conflict) for the sentiment classification task, as analyzed in section \ref{sec:sentiment_analysis_analysis}. 
\begin{table}[htbp]
\centering
\begin{tabular}{ccccc}
\hline
\textbf{Task} & \textbf{LFs Vars} & \textbf{Cov} & \textbf{Over} & \textbf{Conf}\\
\hline
Sentiment & \textbf{v0} & \textbf{21.03} & \textbf{18.21} & \textbf{6.72} \\
 & v1 & 10.63 & 11.37 & 4.41\\\hline
Tags & - & 12.35 & - & - \\\hline 
\end{tabular}
\caption{ The performance of the labeling functions varied for each task(sentiment and tags classification). Sentiment classification labeling functions were evaluated based on coverage, overlaps, and conflict statistics. v0: prioritizing coverage score; v1: prioritizing less conflict. While tags classification labeling functions only evaluated on coverage score, which represented the density of tags}
\label{tab:lf_analysis}
\end{table}
\graphicspath{{images/}}
\begin{figure*}[h]
\centering
\begin{minipage}[b]{0.5\textwidth}
    \includegraphics[width=\textwidth]{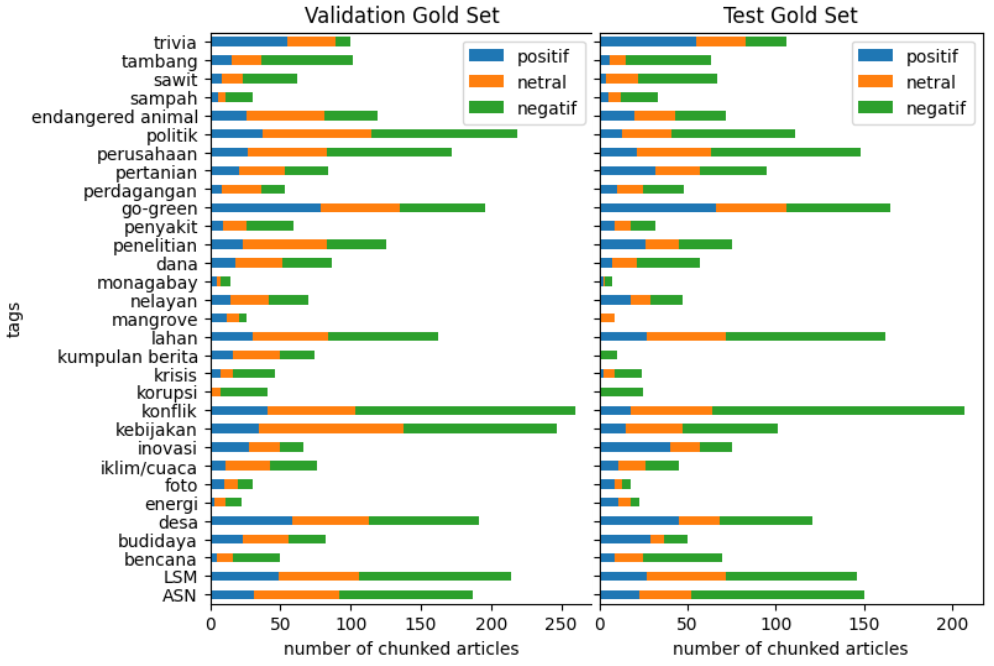}
\end{minipage}
\begin{minipage}[b]{0.3\textwidth}
    \raggedright
    \includegraphics[width=\textwidth]{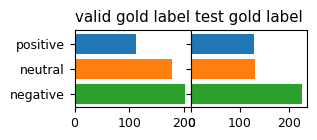}
\end{minipage}
\caption{Distribution of validation and test set gold label for tags classification (left) and sentiment classification (right). Detailed explanation about tags definition was provided in Appendix \ref{sec:tags_explanation}}
\label{fig:data_dist}
\end{figure*}
\subsubsection{Label Model and Final Prediction}
We employed two types of label models in this study: the Covariance Matrix-based label model, adapted from Snorkel \cite{Ratner2016DatProgramming}, and the Majority Voter. The main difference between them lies in their process. The Covariance Matrix requires initial training to generate the final labels using learned weights, whereas the majority voter directly predicts the final labels based on the distribution of labeling function predictions. The hyperparameters used in training Covariance Matrix label model will be provided in Table \ref{tab:label_model_hp}

\section{Experiments}
\subsection{Constructed Dataset}
\label{sec:constructed_dataset}
From the gold-label distribution (Figure \ref{fig:data_dist}), it can be inferred that the sampled Mongabay articles have a bias towards negative sentiment, which is distributed to the entire tags. Refer to (fig \ref{fig:tags_structure}), all tags from \textbf{conflict} are predominantly associated with negative articles, while a few tags from \textbf{trivia} and \textbf{solution} are more commonly found in positive articles, and other tags from \textbf{conflict and solution} were majorly included in neutral articles. It is worth noting that, each article can match with more than one tag considering the varies of author's writing style. \textbf{For example:} one article with negative sentiment can have various tags such as \textit{go-green}, \textit{konflik}, \textit{korupsi}, \textit{LSM}, which indicates the article discusses conflicts in go-green action/regulation, highlights corruption issues, and mentions the involvement of LSM (NGOs). \\
The format of the experimental dataset for training, validation, and testing can be seen in Table \ref{tab:dataset_sample}. Additionally, the constructed dataset, as well as the raw dataset, can be accessed in \url{https://huggingface.co/datasets/Datasaur/mongabay-experiment}.

\begin{table*}
\centering
    \begin{tabular}{l|l|l|l}
    \hline \textbf{Input} & \multicolumn{3}{c}{\textbf{Soft-label}} \\\hline
    Pandemi, Momentum ... | [CLS] Hutan adalah gudang... & \multicolumn{3}{c}{[0.108, 0.001, 0.001, ...} \\\hline \hline
    \hline \textbf{Input} & \multicolumn{3}{c}{\textbf{Soft-label}} \\\hline
    Pandemi, Momentum ... | [CLS] Hutan adalah gudang... & \multicolumn{3}{c}{[1.0, 1.44e-09, 1.32e-09]} \\\hline \hline \hline
    \textbf{Input} & \textbf{ASN} & \textbf{Desa} & \textbf{Konflik}\\ \hline
    Begini Nasib ... | [CLS] Maliau ingin ... & NaN & True & True \\\hline \hline
    \hline \textbf{Input} & \multicolumn{3}{c}{\textbf{Label}} \\\hline
    Begini Nasib ... | [CLS] Maliau ingin ... & \multicolumn{3}{c}{negative} \\\hline    
    \end{tabular}
\caption{Snippet of constructed dataset (top-bottom): \textbf{1:} train set sample with 31 probabilities as soft-label for tags classification experiment \textbf{2:} train set sample with 3 probabilities as soft-label for sentiment classification experiment \textbf{3:} validation set sample with 31 tags in binary label for tags classification experiment \textbf{4:} validation set sample with single-class for sentiment classification experiment}
\label{tab:dataset_sample}
\end{table*}

\subsection{Pipelines}
\label{sec:pipelines}

The experiment pipelines have two main goals: 1) To compare the performance of the covariance matrix and the majority voter-generated dataset for each task, and 2) To assess the performance of different BERT pre-trained models with varying language bases when fine-tuned using the weak dataset. This comparison enables an evaluation of the effectiveness of different approaches to weak supervision for various NLP tasks. \\
To accomplish these objectives, the experiment pipeline is structured as follows:
\begin{itemize}
    \item Each variation of the pre-trained models will be trained using the soft-label dataset (both the covariance matrix and the majority voter version).
    \item The models will be iteratively evaluated using the gold-label validation set at each epoch.
    \item The weight configuration yielding the best validation metrics will be tested using the gold-label test set.
\end{itemize}
This pipeline was executed for both the tags classification and sentiment classification tasks.

\subsection{Finetuning}
\label{sec:finetuning}
We utilized various BERT variations from HuggingFace's pretrained language models: \verb|indobert-base-uncased|\footnote{\url{https://huggingface.co/indolem/indobert-base-uncased}}, \verb|bert-base-multilingual-cased|\footnote{\url{https://huggingface.co/bert-base-multilingual-cased}}, and \verb|bert-base-cased|\footnote{\url{https://huggingface.co/bert-base-cased}}. We employed these models to assess the performance of our weakly labeled dataset when learned by three distinct models: 1) Indonesian monolingual (pre-trained with the same language as our data), 2) Multilingual (pre-trained with multiple languages, including the same language as our data), 3) English monolingual (pre-trained with a language different from our data). \\
In finetuning hashtag classification, we utilized cross-entropy loss instead of binary-cross-entropy loss. However, in the inference session, we keep using binary-cross-entropy loss. This decision is based on the soft-label distribution of the training set obtained from the weak supervision process, which is not binary for each class, in contrast to the binary distribution of the gold labels in the validation and test sets. We utilize the hyperparameter setup outlined in (Table \ref{tab:finetuning_hp}) for the fine-tuning of each task.

\begin{table*}
\centering
\begin{tabular}{llllll}
\hline
\textbf{PreModel} & \textbf{Aggr} & \textbf{Acc (Val)} & \textbf{F1 (Val)} & \textbf{Acc 
 (Test)} & \textbf{F1 (Test)} \\ \hline
\verb|indobert| & CM & 60.77 & 56.37 & 59.79 & 55.72 \\
& MV & 58.74 & 53.97 & 58.97 & 54.12 \\ \hline
\verb|mbert| & CM & 55.08 & 50.25 & 50.93 & 44.16 \\
& MV & 55.89 & 42.45 & 49.9 & 38.25 \\ \hline
\verb|bert| & CM & 46.16 & 36.43 & 44.74 & 35.83 \\
& MV & 44.51 & 34.55 & 42.68 & 33.22 \\ \hline
\end{tabular}
\caption{Validation and test results from sentiment classification experiment, use labeling functions v0. The performance was gained from a model with the best validation score. \textbf{CM}: using Covariance Matrix as label model; \textbf{MV}: using Majority Voter as label model}
\label{tab:sentiment_exp_result}
\end{table*}
\begin{table*}
\centering
\begin{tabular}{cccccccc}
\hline
\textbf{LFs var} & \textbf{cov} & \textbf{over} & \textbf{conf} & \textbf{ACC (Val)} & \textbf{F1 (Val)} & \textbf{ACC test} & \textbf{F1 test} \\ \hline
v0 & 21.03 & 18.21 & 6.72 & 60.77 & 56.37 & 59.79 & 55.72 \\
v1 & 6.9 & 2.31 & 0.21 & 56.71 & 45.27 & 50.31 & 40.49 \\ \hline
\end{tabular}
\caption{Validation and test results from sentiment classification labeling function variations experiment using indobert and Covariance Matrix. High coverage (v0), even with high conflict, gives the best performance than more precise and accurate labeling functions result (v1)}
\label{tab:lf_exp_result}
\end{table*}

\subsection{Analysis}
\label{sec:general_analysis}
The experiment results were analyzed from two key perspectives: 1) Label model comparison perspective, and 2) Dataset learnability.\\
From the perspective of dataset learnability, the Learning process went well by the Indobert pre-trained model, represented by significantly better performance compared to the Multilingual BERT (mbert) and BERT Base models, as evident in the results presented in Table \ref{tab:sentiment_exp_result}, Figure \ref{fig:loss_sentiment}, and Table \ref{tab:tags_exp_result}\\
In the context of label model comparison, the results vary between multi-label and sentiment classification tasks. For sentiment classification (as shown in Table \ref{tab:sentiment_exp_result}), the Covariance Matrix (CM) approximates the human judgment (gold label) more accurately. In contrast, Majority Voter (MV) predictions are more closely correlated with the gold label in multi-label classification, as indicated in Table \ref{tab:tags_exp_result}.

\subsubsection{Sentiment Classification}
\label{sec:sentiment_analysis_analysis}
\begin{figure}[h]
    \centering
    \includegraphics[width=7.5cm]{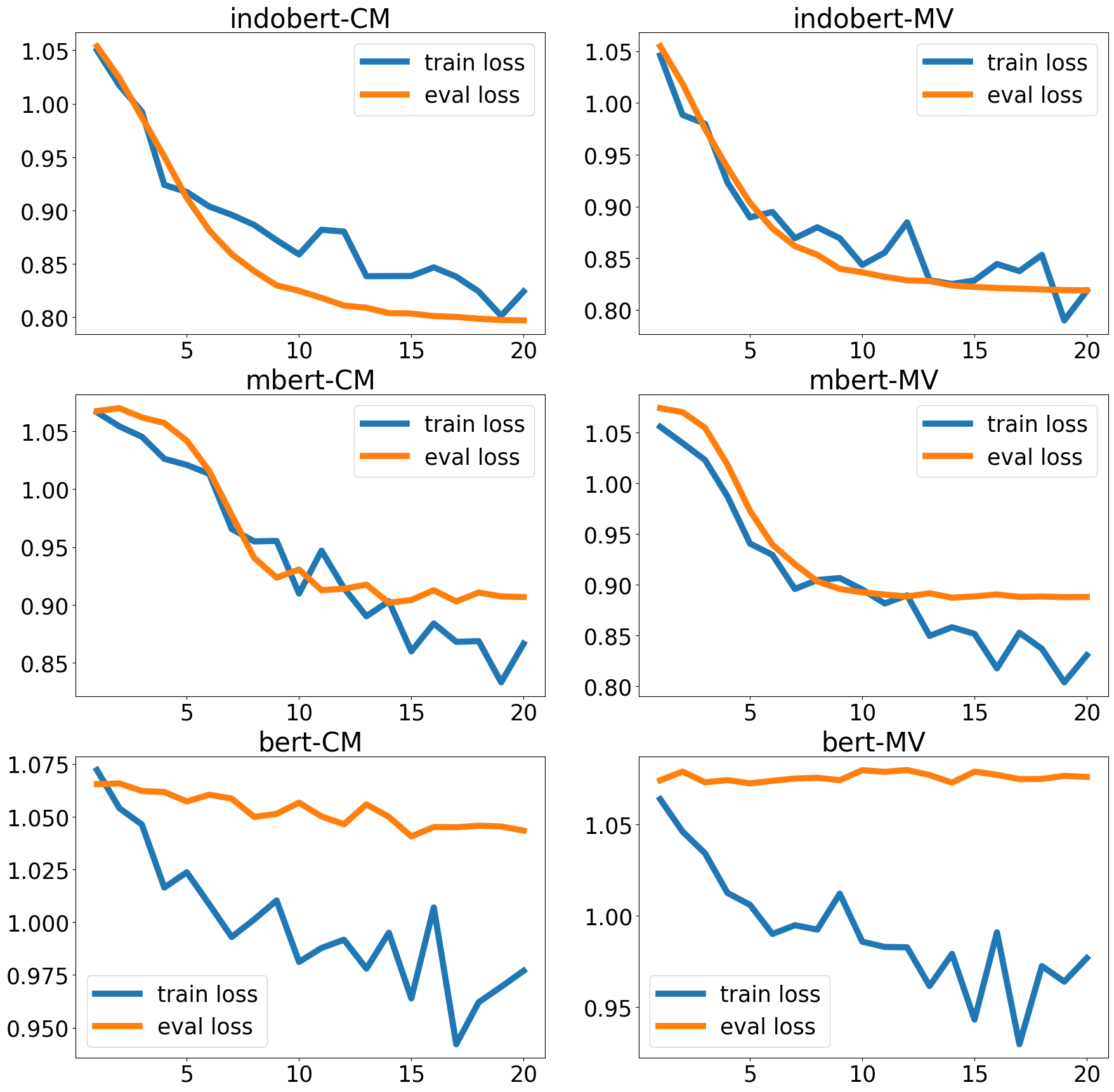}
    \caption{Train-eval loss graph of sentiment classification finetuning on 3 pre-trained models and 2 label models. \textbf{Left:} Loss graph of CovarianceMatrix across indobert, multilingual-bert (mbert), bert-base (bert) (top-bottom). \textbf{Right:} Loss graph of MajorityVoter across indobert, multilingual-bert (mbert), bert-base (bert) (top-bottom)}
    \label{fig:loss_sentiment}
\end{figure}

The highest performance, reaching 60\% accuracy and a 56\% F1-score (macro average), was achieved by utilizing the Covariance Matrix label model in conjunction with the Indobert pre-trained model, as shown in Table~\ref{tab:sentiment_exp_result}. \\
Detailed sentiment classification F1-scores for each label (refer to Appendix \ref{sec:f1_score_label}) reveal that our fine-tuned models (Indobert and Multilingual-Bert) exhibit a tendency to predict negative articles accurately, with F1-scores exceeding 70\%. This observation aligns with our dataset's characteristic (Figure \ref{fig:data_dist}), where negative articles are prominently distributed. However, Indobert faces challenges when predicting articles associated with the positive or neutral class, resulting in F1-scores hovering around 40\%. This could be attributed to the similarity between positive and neutral-related tags (Figure \ref{fig:data_dist}). \\
Indobert also exhibits a faster learnability rate compared to other pretrained models, supported by the loss graph in Figure \ref{fig:loss_sentiment}. The comparison between training and evaluation losses indicates that Indobert's loss fits well, while Multilingual-Bert displays signs of overfitting around the 15th epoch, and Bert-Base encounters difficulties in learning the dataset from the outset.
\graphicspath{{images/}}
\begin{table*}[h]
\centering
\begin{tabular}{llllllll}
\hline
\textbf{PreModel} & \textbf{Aggr} & \textbf{R/A (Val)} & \textbf{F1-ma (Val)} & \textbf{F1-mi (Val)} & \textbf{R/A 
 (Test)} & \textbf{F1-ma (Test)} & \textbf{F1-mi (Test)} \\ \hline
\verb|indobert| & CM & 82.93 & 73.33 & 78.3 & 81.89 & 65.71 & 69.9\\
& MV & 85.61 & 76.39 & 81.7 & 83.67 & 66.87 & 71.5\\ \hline
\verb|mbert| & CM & 81.16 & 69.86 & 75 & 80.01 & 62.18 & 66.8\\
& MV & 85.14 & 75.49 & 80.7 & 82.85 & 64.95 & 70.2\\ \hline
\verb|bert| & CM & 70.07 & 47.07 & 55 & 69.56 & 44.45 & 49.2\\
& MV & 73.34 & 53.25 & 61.23 & 73.16 & 51.96 & 55.4\\ \hline
\end{tabular}
\caption{Validation and test results from tags classification experiment. The performance was gained from a model with the best validation score. \textbf{CM}: Using Covariance Matrix as label model; \textbf{MV}: Using Majority Voter as label model. \textbf{R/A}:ROC-AUC; \textbf{F1-ma}: F1-score macro average; \textbf{F1-mi}: F1-score micro average}
\label{tab:tags_exp_result}
\end{table*}
Additionally, our supplementary experiment explores how the quality of labeling functions affects model performance. As shown in Table \ref{tab:lf_exp_result}, despite LFs v0 having ten times higher conflict proportion compared to LFs v1, the v0 labeling functions significantly outperform v1 in fine-tuned models. This suggests that higher coverage and overlaps contribute to enhanced performance in fine-tuned models, even in the presence of increased conflict. This finding is consistent with the statement in \citet{Ratner:2020} that higher coverage results in higher accuracy.

\subsubsection{Tags Classification}
\label{sec:tags_classification}
The tags classification results are depicted in micro-average F1-score and ROC-AUC due to the imbalanced distribution of tags in our dataset (Figure \ref{fig:data_dist}). \\ 
As stated earlier in Section \ref{sec:general_analysis}, Majority Voter (MV) consistently outperforms the Covariance Matrix (CM) by approximately 2-3\%. This aligns with the findings in \citet{zhang2021wrench}, which reported that Majority Voters as the label model achieved superior performance when dealing with sparse labels in tags classification. The highest performance is observed when using Majority Voter as the label model and Indobert for finetuning, resulting in a test performance of 81.89\% ROC-AUC, 65.71\% F1-score$_{macro\_average}$, and 69.9\% F1-score$_{micro\_average}$. \\
A more detailed analysis of tags classification F1-scores for each label is provided in Appendix \ref{sec:f1_score_label}. Tags associated with negative sentiment achieve F1-scores exceeding 70\% with either Indobert or Multilingual-Bert. In contrast, tags primarily composed of non-negative sentiment classes do not perform as effectively. This result aligns with the sentiment classification results. \\
Since we utilize different losses during training and inference sessions, we cannot compare the train and eval loss in a single frame (Section \ref{sec:finetuning}).

\section{Limitations}
Our dataset construction using weak supervision relies heavily on our labelers' subjectivity, particularly in collecting hashtags and creating labeling functions. These functions are designed to match the unique characteristics of our dataset, and their effectiveness may not extend well to other datasets, even with similar characteristics. The low results in our experiments are attributed to biases within the dataset. However, these biases provide insights into the Indonesian environmental and conservation editorial landscape, albeit presenting challenges for future efforts in creating more equitable datasets.\\
Although our data programming is primarily designed for single/multi-class classification, through this work it can be utilized for multi-label classification due to the one-hot-encoded output format. However, we have not yet implemented any metrics for evaluating labeling function performance, aside from coverage, which represents the dataset's labeling density.

\section{Conclusion}
In conclusion, we've presented our weak supervision pipeline for creating two datasets: sentiment classification and multi-label classification. We utilized Mongabay conservation articles collection to construct our datasets and adapted Snorkel's framework in Datasaur's workspace for conducting this work. Our results show that utilizing Data Programming to curate datasets can deliver qualified datasets that are learnable by BERT pre-trained models, especially indobert. However, some limitations remain, such as labeling function subjectivity, incomplete multi-label classification labeling function metrics, and the implicit bias within this dataset. Future work will curate more datasets more robustly and reproducibly, especially for NLP datasets with underrepresented language, topic, or task. 

\section*{Ethics Statement}
 We acknowledge the presence of implicit biases in both our dataset source and the constructed dataset. Additionally, as we utilized news datasets, they may contain certain viewpoints from editors or journalists. It is ensured that the dataset is free from harmful or offensive content, biases may still exist in our model and results. As a news dataset, it contains formal Indonesian-native content.
 
\section*{Acknowledgements}
We extend our heartfelt gratitude to Datasaur.ai\footnote{\url{https://datasaur.ai/}} for their generous financial and resource support, which was instrumental in making this research possible. Their assistance provided us with essential resources, especially computational support, and allowed us to collaborate with experts in the field. We sincerely appreciate the entire team at Datasaur.ai for their invaluable support and partnership throughout this study.

We would also thank Mongabay\footnote{\url{https://www.mongabay.co.id/}} conservation portal for providing the dataset used in this study. The extensive collection of articles rich in Indonesian environmental subjects, issues, and concerns really helps in curating our Indonesian NLP dataset based on real-world conditions.

\bibliography{main}
\bibliographystyle{acl_natbib}

\onecolumn
\appendix

\section{Appendix}
\subsection{Labeling Function Construction}
\label{sec:lf_appendix}

Our labeling functions were supported with external Python libraries, such as SpaCy\footnote{\url{https://spacy.io/}}, NLTK\footnote{\url{https://www.nltk.org/}}, TextBlob\footnote{\url{https://textblob.readthedocs.io/en/dev/}}, and Stanza\footnote{\url{https://stanfordnlp.github.io/stanza/}}. We have standardized the labeling functions' code into this template. So varies of Python algorithms can be implemented under \texttt{label\_function}.

\definecolor{backcolour}{rgb}{0.95,0.95,0.92}

\lstdefinestyle{mystyle}{
    backgroundcolor=\color{backcolour},
    keywordstyle=\color{magenta},
    numberstyle=\tiny\color{codegray},
    basicstyle=\ttfamily\scriptsize,
    breakatwhitespace=false,         
    breaklines=true,                 
    captionpos=b,                    
    keepspaces=true,                 
    numbersep=5pt,                  
    showspaces=false,                
    showstringspaces=false,
    showtabs=false,                  
    tabsize=1
}

\lstset{style=mystyle}

\begin{lstlisting}[language=Python]
LABELS = {
  'labelA' : 0,
  'labelB' : 1,
  'labelC' : 2
}


# decorator to wrap labeling function into ready output 
# to be aggregated by label model

#assign target label
@target_label(label=LABELS['labelA'])
def label_function(sample) -> bool:
  #this line created text with all columns of table
  #dataset in this case title and article columns
  text = list(sample.values())[0]

  # Implement your logic here
  TARGET_KEYWORDS = ['special keywords']
  for keyword in TARGET_KEYWORDS:
    if re.search(keyword, text, re.IGNORECASE):
      return True
  return False
\end{lstlisting}

For the simplest labeling functions, we only inserted keywords into \verb|TARGET_KEYWORDS|, and add additional code/rules below \verb|LABELS| when required.

\subsubsection{Tag Classification}

\textit{\textbf{targeted tags:} pertanian, penelitian, lembaga internasional, berita internasional, Aparatur Sipil Negara, kebijakan, hewan terancam punah, bencana alam, konflik, tambang, perusahaan, pendanaan, penyelamatan lingkungan, sawit, lahan, Lembaga Swadaya Masyarakat, foto, trivia, mangrove, politik, masyarakat desa, inovasi, tentang mongabay, perdagangan, kumpulan berita, krisis, penyakit, sampah, korupsi, budidaya, nelayan, iklim/cuaca, energi}
\\\\
To address the 31 tags, we developed a total of 36 labeling functions, categorized based on their complexity. The least complex labeling functions applied rules directly to the data point, which consisted of the \verb|title| and \verb|chunked-article|. Medium complexity labeling functions involved separate keyword searches in the \verb|title| and \verb|chunked-article|. The most complex labeling functions required more intricate logic and rules.

\begin{itemize}
    \item The simplest labeling functions consists of rules that return tags labels related to obvious keywords. These tags represent specific topics and can be identified by a single or a few keywords.
    
\begin{lstlisting}
budidaya= ['perudangan', 'pembenihan', 'budidaya', 'tambak', 'keramba', 'budi daya', 'perikanan']
energi= ['Energi', 'energi', 'bahan bakar,', 'BBM', 'batubara', 'batu bara']
foto= ['Foto', 'foto utama']
korupsi= ['korupsi', 'koruptor', 'korup']
krisis= ['krisis']
kumpulan berita= ['catatan awal tahun', 'catatan akhir tahun', 'kaleidoskop']
mangrove= ['mangrove']
nelayan= ['nelayan', 'melaut']
perdagangan= ['pemesan', 'perdagangan', 'ekspor', 'export', 'impor', 'trade', 'market']
pertanian= ['pertanian', 'bertani', 'sawah', 'ladang', 'pupuk','pestisida']
endangered animal= ['predator', 'diburu', 'populasi', 'habitat', 'peliharaan', 'penangkaran', 'conservation', 'habitat', 'pelepasliaran', 'hutan lindung', 'langka ', 'punah', 'endangered', 'cagar alam', 'suaka margasatwa', 'kebun binatang']
sampah= ['sampah', 'cemari', 'tercemar']
sawit= ['sawit']
tambang= ['tambang', 'nambang']
\end{lstlisting}
    \item The medium labeling functions utilize common additional logic, such as searching for keywords in separate parts (\verb|title| and \verb|chunked-article|) and combining them, searching for keywords based on the case or uncased first character, or employing simple regex patterns for keyword matching.\\
\begin{lstlisting}[language=Python, caption=ASN]
def label_function(sample) -> bool:
  text = sample['title']+' | '+sample['text']

  # Implement your logic here
  TARGET_KEYWORDS_UNCASED = ['jendral', 'sekretariat', 'pasukan', 'komnas', 'dirjen', 'direktorat', 'mahkamah', 'militer', 'aparat', 'polda', 'polres', 'polsek', 'polri', 'kepala desa', ' dinas', 'polisi']
  TARGET_KEYWORDS_CASED = ['BPD', 'KPK', 'MA', 'MK', 'PNS', 'TNI']
  if any(re.search(keyword, text) for keyword in TARGET_KEYWORDS_CASED) or \
  any(re.search(keyword, text, re.IGNORECASE) for keyword in TARGET_KEYWORDS_UNCASED):
    return True
  return False
\end{lstlisting}
\begin{lstlisting}[language=Python, caption=LSM]
def label_function(sample) -> bool:
  text = sample['title']+' | '+sample['text']

  # Implement your logic here
  TARGET_KEYWORDS = ['Alliance', 'Liga', 'Duta', 'Yayasan', 'LSM', 'Balai', 'Society', 'Lembaga', 'Perhimpunan', 'WWF-Indonesia', 'Greenpeace', 'Friends', 'Badan', 'Community', 'Komunitas', 'Forum', 'Dewan', 'Support', 'Lovers', 'PKK']
  for keyword in TARGET_KEYWORDS:
    if re.search(keyword, text):
      if 'DPRD' in text or 'Dewan Perwakilan Rakyat' in text:
        continue
      return True
  return False
\end{lstlisting}
\begin{lstlisting}[language=Python, caption=bencana]
def label_function(sample) -> bool:
  title = sample['title']
  text = title+' | '+sample['text']

  # Implement your logic here
  TARGET_KEYWORDS_TITLE = ['Bencana', 'Mitigasi']
  TARGET_KEYWORDS = ['kebakaran', 'banjir', 'gempa', 'tsunami', 'erupsi', 'longsor']
  if any(re.search(keyword,title) for keyword in TARGET_KEYWORDS_TITLE) or \
  any(re.search(keyword, text, re.IGNORECASE) for keyword in TARGET_KEYWORDS):
    return True
  return False
\end{lstlisting}
\begin{lstlisting}[language=Python, caption=inovasi]
def label_function(sample) -> bool:
  title = sample['title']
  text = sample['text']

  # Implement your logic here
  TARGET_KEYWORDS_TITLE = ['Penelitian', 'Riset', 'Studi', 'Analisis', 'Sukses', 'Inovasi', 'Alternatif', 'Manfaat', 'Sains', 'Potensi', 'Asa ', 'Solusi']
  EX_KEYWORDS_TITLE = ['Tapi', 'Tidak']
  TARGET_KEYWORDS_TEXT = ['sukses', 'inisiatif', 'penghargaan']
  if any(re.search(keyword,title) for keyword in TARGET_KEYWORDS_TITLE) and all(re.search(ex_keyword, title)==None for ex_keyword in EX_KEYWORDS_TITLE) or\
    any(re.search(keyword, text, re.IGNORECASE) for keyword in TARGET_KEYWORDS_TEXT):
      return True
  return False
\end{lstlisting}
\begin{lstlisting}[language=Python, caption=kebijakan]
def label_function(sample) -> bool:
  title = sample['title']
  text = title+' | '+sample['text']

  # Implement your logic here
  TARGET_KEYWORDS_TITLE = ['Kebijakan']
  TARGET_KEYWORDS_CASED = ['UU', 'SK', 'PP', 'REDD', 'Permen']
  TARGET_KEYWORDS_UNCASED = ['zonasi', ' perda ', 'perizinan', 'moratorium', 'lisensi', 'undang', 'legalitas', 'revisi', 'regulasi', 'restorasi']
  if any(re.search(keyword,title) for keyword in TARGET_KEYWORDS_TITLE) or \
  any(re.search(keyword, text) for keyword in TARGET_KEYWORDS_CASED) or \
  any(re.search(keyword, text, re.IGNORECASE) for keyword in TARGET_KEYWORDS_UNCASED):
    return True
  return False
\end{lstlisting}
\begin{lstlisting}[language=Python, caption=konflik]
def label_function(sample) -> bool:
  title = sample['title']
  text = sample['title']+' | '+sample['text']

  # Implement your logic here
  TARGET_KEYWORDS_TITLE = ['Mangkir', 'Merana', 'Diburu', 'Merusak', 'Kesulitan', 'Disita', 'Selundupkan', 'Memusuhi', 'Rumit', 'Perusak', 'Terpaksa', 'Kurang Gizi', 'Protes', 'Kehilangan', 'Nestapa', 'Ilegal', 'Kriminal', 'Balada', 'Lapor ', 'Terancam', 'Usir', 'Musuh', 'Terkontaminasi', 'Kasus']
  TARGET_KEYWORDS = ['pelanggaran hukum', 'pemaksaan', 'beracun', 'melanggar', 'kebakaran', 'muak','egois', 'over populasi', 'perusakan', 'membahayakan', 'kriminal', 'serakah', 'kejam', 'kelaparan', 'tak becus', 'eksploitasi', 'pengeboman', 'penjahat', 'diracun', 'penyuapan', 'teror', 'intimidasi', 'illegal', 'ilegal', 'polemik', 'tercemar', 'cemar', 'modus', 'diabaikan', ' tambang', 'korupsi', 'koruptor', 'korup', 'nyiksa', 'siksa', 'penjara', 'aniaya', 'diancam', 'mengancam', 'pembunuh', 'dibunuh', 'membunuh', 'terbunuh', 'tembak', 'nembak', 'miris', 'konflik', 'bentrok', 'tuduh', 'tuding', 'tuntut', 'nuntut']
  EX_KEYWORDS_TITLE = ['Tidak Berbahaya', 'Paling', 'Konservasi', 'Ini ', 'Menentramkan', 'Karya', 'Festival', 'Ke Alam', 'Menghijaukan', 'DNA', 'Perjuangan', 'Kerjasama', 'unik', 'Belajar', 'Dilepasliarkan', 'Tidak Berbahaya', 'Manfaat', 'Ayo', 'Solusi', 'Restorasi', 'Inilah', 'Sayang', 'Transparansi', 'Keren', 'Mengubah Dunia', 'Konsolidasi', 'Diselamatkan', ' Si ', 'Berjuang', 'Evolusi', 'Foto:', 'Penyelamatan', 'Disahkan', 'Bantuan', 'Mengakhiri Penebangan Liar', 'Berhasil', 'Bersihkan', 'Penjaga', 'Perkuat', 'Aplikasi', 'Sukses', 'Anti', 'Perlindungan', 'Agen Rahasia', 'Warrior']
  EX_KEYWORDS = ['kerusakan', 'sustainable', 'teknologi canggih', 'rehabilitasi', 'bersinergi', 'konservasi', 'presisi', 'untunglah', 'komitmen', 'edukasi', 'kredibilitas', 'citation', 'reward', 'penghargaan', 'inisiatif', 'pulih']
  if (any(re.search(keyword, title) for keyword in TARGET_KEYWORDS_TITLE) or \
  any(re.search(keyword, text, re.IGNORECASE) for keyword in TARGET_KEYWORDS)) and \
  (all(ex_keyword not in title for ex_keyword in EX_KEYWORDS_TITLE) and \
  all(ex_keyword not in text for ex_keyword in EX_KEYWORDS)):
    return True
  return False
\end{lstlisting}
\begin{lstlisting}[language=Python, caption=lahan]
def label_function(sample) -> bool:
  title = sample['title']
  text = sample['text']

  # Implement your logic here
  TARGET_KEYWORDS_TITLE = ['tanah']
  TARGET_KEYWORDS_TEXT = ['lahan']
  if any(re.search(keyword,title) for keyword in TARGET_KEYWORDS_TITLE) or\
    any(re.search(keyword, text, re.IGNORECASE) for keyword in TARGET_KEYWORDS_TEXT):
      return True
  return False
\end{lstlisting}
\begin{lstlisting}[language=Python, caption=mongabay]
def label_function(sample) -> bool:
  title = sample['title']
  text = sample['text']

  # Implement your logic here
  TITLE_KEYWORDS = ['Mongabay']
  TEXT_KEYWORDS = ['Mongabay.org']

  if any(re.search(keyword, title) for keyword in TITLE_KEYWORDS) or\
  any(re.search(keyword, text) for keyword in TEXT_KEYWORDS):
    return True
  return False
\end{lstlisting}
\begin{lstlisting}[language=Python, caption=penelitian]
def label_function(sample) -> bool:
  title = sample['title']
  text = sample['text']

  # Implement your logic here
  TARGET_KEYWORDS_TITLE = ['LIPI', 'Ahli', 'Studi', 'Kajian', 'Aplikasi', 'Pakar', 'Peneliti', 'Riset', 'Analisis', 'Analisa', 'Inovasi', 'Alternatif', 'Sains']
  TARGET_KEYWORDS_CASED = ['LIPI', 'M.Sc', 'M.Si', 'CITATION', 'DNA', 'PhD']
  TARGET_KEYWORDS_UNCASED = ['ahli biologi', 'hasil penelitian', 'pakar ', 'peneliti ', 'universitas', 'university', 'profesor', 'professor', 'science', 'ilmuwan', 'scientist', 'research', 'college']
  if any(re.search(keyword,title) for keyword in TARGET_KEYWORDS_TITLE) or \
  any(re.search(keyword, text) for keyword in TARGET_KEYWORDS_CASED) or \
  any(re.search(keyword, text, re.IGNORECASE) for keyword in TARGET_KEYWORDS_UNCASED) or \
  len(re.findall('ilmiah', text, re.IGNORECASE)) > 1:
    return True
  return False
\end{lstlisting}
\begin{lstlisting}[language=Python, caption=penyakit]
def label_function(sample) -> bool:
  title = sample['title'] 
  text = title+' | '+sample['text']

  # Implement your logic here
  TARGET_KEYWORDS_TITLE = ['penyakit', 'virus', 'parasit', 'covid', 'corona']
  TARGET_KEYWORDS = ['infeksi ', 'menular', ' medis ']
  if any(re.search(keyword, text, re.IGNORECASE) for keyword in TARGET_KEYWORDS) or \
  any(re.search(keyword, title, re.IGNORECASE) for keyword in TARGET_KEYWORDS_TITLE):
    return True
  return False
\end{lstlisting}
\begin{lstlisting}[language=Python, caption=perusahaan]
def label_function(sample) -> bool:
  text = sample['title'] +' | '+ sample['text']

  # Implement your logic here
  TARGET_PATTERNS = [re.compile('\bPT '), re.compile('\bPT.'), ' PT ', ' CV ']
  TARGET_KEYWORDS = ['perusahaan','.org', 'direksi', 'badan usaha', 'industri', 'pabrik', 'produsen']

  if any(re.search(keyword, text, re.IGNORECASE) for keyword in TARGET_KEYWORDS) or \
  any(re.search(pattern, text) for pattern in TARGET_PATTERNS):
    return True
  return False
\end{lstlisting}

    \item The most complex labeling functions, handle tags that are related to common keywords that are also found commonly in regular content. In such cases, more detailed and complex rules need to be added in combination with keywords. This type of tags is typically handled by multiple labeling functions and may require the use of external packages, such as spaCy.
\begin{lstlisting}[language=Python, caption=desa-keywords]
def label_function(sample) -> bool:
  text = sample['title']+' | '+sample['text']

  # Implement your logic here
  TARGET_KEYWORDS_CASED = ['Adat', 'Suku']
  TARGET_KEYWORDS_UNCASED = ['desa', 'dusun']
  
  if any(re.search(keyword, text) for keyword in TARGET_KEYWORDS_CASED) or\
  any(re.search(keyword, text, re.IGNORECASE) for keyword in TARGET_KEYWORDS_UNCASED):
    return True
  return False
\end{lstlisting}
\begin{lstlisting}[language=Python, caption=desa-title-masyarakat-logic]
def label_function(sample) -> bool:
  text = list(sample.values())[0]

  # Implement your logic here
  first_word = text.split()[0]
  if first_word == 'Masyarakat':
    return True
  elif 'Warga' in sample['title']:
    return True
  return False
\end{lstlisting}
\begin{lstlisting}[language=Python, caption=iklim/cuaca-keywords]
def label_function(sample) -> bool:
  text = sample['title']+' | '+sample['text']

  # Implement your logic here
  TARGET_KEYWORDS = ['iklim', 'cuaca', 'pemanasan global', 'el-nino', 'el nino', 'siklon']
  for keyword in TARGET_KEYWORDS:
    if re.search(keyword, text, re.IGNORECASE):
      return True
  return False
\end{lstlisting}
\begin{lstlisting}[language=Python, caption=iklim/cuaca-repetitive keywords]
def label_function(sample) -> bool:
  text = sample['title']+' | '+sample['text']

  # Implement your logic here
  TARGET_KEYWORDS = ['badai', 'kemarau', 'hujan']
  match_collection = []
  for keyword in TARGET_KEYWORDS:
    match_collection.extend(re.findall(keyword, text, re.IGNORECASE))
      
  if len(match_collection)>3:
    return True
  else:
    return False
\end{lstlisting}
\begin{lstlisting}[language=Python, caption=pendanaan-keywords-regex number]
def label_function(sample) -> bool:
  text = sample['title']+' | '+sample['text']

  TARGET_PATTERNS = []
  nominal_keywords = ["juta","miliar","triliun"]
  for nominal in nominal_keywords:
    TARGET_PATTERNS.append(re.compile(r"Rp\d+[,|.]\d+ {}".format(nominal)))
    TARGET_PATTERNS.append(re.compile(r"Rp \d+[,|.]\d+ {}".format(nominal)))
    TARGET_PATTERNS.append(re.compile(r"Rp \d+ {}".format(nominal)))
    TARGET_PATTERNS.append(re.compile(r"Rp\d+ {}".format(nominal)))
    
  rp_2 = re.compile(r"Rp \d+.\d+")
  rp_3 = re.compile(r'Rp\d+.\d+')
  TARGET_PATTERNS.append(rp_2)
  TARGET_PATTERNS.append(rp_3)

  nominal_keywords = ["puluh","ratus","ribu","juta","miliar"]
  for nominal in nominal_keywords:
    TARGET_PATTERNS.append(re.compile(r"\d+ {} dollar [AS|Australia]".format(nominal)))
    TARGET_PATTERNS.append(re.compile(r"\d+ {} dollar".format(nominal)))
    TARGET_PATTERNS.append(re.compile(r"\$\d+[,|.]\d+ {}".format(nominal)))
    TARGET_PATTERNS.append(re.compile(r"\$ \d+[,|.]\d+ {}".format(nominal)))
    TARGET_PATTERNS.append(re.compile(r"\$ \d+ {}".format(nominal)))
    TARGET_PATTERNS.append(re.compile(r"\$\d+ {}".format(nominal)))
    
  dollar_2 = re.compile(r"\$ \d+.\d+")
  dollar_3 = re.compile(r'\$\d+.\d+')

  TARGET_PATTERNS.append(dollar_2)
  TARGET_PATTERNS.append(dollar_3)  

  # Implement your logic here
  TARGET_KEYWORDS = [' dana', 'pendanaan', 'dollar', 'rupiah']

  if any(re.search(keyword, text, re.IGNORECASE) for keyword in TARGET_KEYWORDS) or \
  any(re.search(pattern, text) for pattern in TARGET_PATTERNS):
    return True
  return False
\end{lstlisting}
\begin{lstlisting}[language=Python, caption=go green (penyelamatan lingkungan) - keywords]
def label_function(sample) -> bool:
  text = sample['title']+' | '+sample['text']

  # Implement your logic here
  TARGET_KEYWORDS = ['pelestarian lingkungan', 'revitalisasi', 'melindungi hutan', 'kampanye lingkungan', 'kontribusi', 'gerakan', 'ramah lingkungan', 'pegiat']
  
  for keyword in TARGET_KEYWORDS:
    if re.search(keyword, text, re.IGNORECASE):
      return True
  return False
\end{lstlisting}
\begin{lstlisting}[language=Python, caption=go green (penyelamatan lingkungan) - keywords - logic]
def label_function(sample) -> bool:
  text = sample['title']

  # Implement your logic here
  TARGET_KEYWORDS = ['pelestarian', 'mengelola', 'merawat alam', 'pengelolaan', 'penghargaan', 'aksi', 'hijau', 'kisah', 'jaga', 'selamat', 'menyelamatkan', 'penyelamat', 'aktivis', 'karya', 'ayo', 'mengajak', 'anti-illegal', 'warior']
  EX_KEYWORDS = ['penyu', 'bunglon']
  for keyword in TARGET_KEYWORDS:
    if re.search('Ajak', text):
      return True
    elif re.search(keyword, text, re.IGNORECASE) and re.search('konflik', sample['text'], re.IGNORECASE)==None:
      if keyword == 'hijau':
        try:
          previous_text = text[:text.index(keyword)]
          if any(re.search(ex_keyword,previous_text,re.IGNORECASE) for ex_keyword in EX_KEYWORDS):
            continue
        except:
          pass
      return True
  return False
\end{lstlisting}
\begin{lstlisting}[language=Python, caption=politic-keywords]
def label_function(sample) -> bool:
  title = sample['title']
  text = sample['title']+' | '+sample['text']

  # Implement your logic here
  TARGET_KEYWORDS_TITLE = ['Presiden', 'Politik', 'Pemerintah']
  TARGET_KEYWORDS_CASED = ['DPR', 'MPR', 'KPK']
  TARGET_KEYWORDS_UNCASED = ['partai', 'pemilu', 'capres', 'cawapres', 'presiden indonesia', 'wali kota', 'walikota', 'bupati', 'gubernur', 'menteri', 'mentri', 'pemkot', 'pemda', 'pemprov', 'pejabat', 'parpol', 'pilkada', 'pilpres', 'politis', 'politikus']
  
  if any(re.search(keyword, text) for keyword in TARGET_KEYWORDS_CASED) or\
  any(re.search(keyword, text, re.IGNORECASE) for keyword in TARGET_KEYWORDS_UNCASED) or\
  any(re.search(keyword, title) for keyword in TARGET_KEYWORDS_TITLE):
    return True
  
  return False
\end{lstlisting}
\begin{lstlisting}[language=Python, caption=politic-text-pemerintah-logic]
def label_function(sample) -> bool:
  text = sample['text']

  # Implement your logic here
  TARGET_KEYWORDS = ['pemerintah']
  match_collection = []
  for keyword in TARGET_KEYWORDS:
    match_collection.extend(re.findall(keyword, text, re.IGNORECASE))
  
  if len(match_collection)>3:
    return True
  return False
\end{lstlisting}
\begin{lstlisting}[language=Python, caption=trivia-keywords]
def label_function(sample) -> bool:
  text = sample['text']

  # Implement your logic here
  TARGET_KEYWORDS = ['pemerintah']
  match_collection = []
  for keyword in TARGET_KEYWORDS:
    match_collection.extend(re.findall(keyword, text, re.IGNORECASE))
  
  if len(match_collection)>3:
    return True
  return False
\end{lstlisting}
\begin{lstlisting}[language=Python, caption=trivia-title-PER-spaCy]
def label_function(sample) -> bool:
  text = sample['title']

  try:
    translated_text = TextBlob(text).translate(from_lang='id', to='en')
  except:
    translated_text = TextBlob(text)

  trans_text_split = str(translated_text).split(',')
  if len(trans_text_split) == 1:
    trans_text_split = str(translated_text).split(':')

  if len(trans_text_split) > 1:
    name_candidate = trans_text_split[0]
    
    spacy_pred = nlp(name_candidate)
    TARGET_KEYWORDS = []
    for token in spacy_pred.ents:
      token_label = token.label_
      if token_label in NER_LABELS:
        TARGET_KEYWORDS.append(str(token))

    # Implement your logic here
    for keyword in TARGET_KEYWORDS:
      if re.search(keyword, text, re.IGNORECASE):
        return True
  return False
\end{lstlisting}
\end{itemize}

\subsubsection{Sentiment Classification}
The main idea of sentiment classification labeling functions is to gather groups of tags that are related to specific sentiment classes (positive, neutral, or negative). However, we discovered that labeling functions based solely on tags often resulted in conflicts between different labeling functions. To address this issue, we created two versions of labeling functions: one that relies solely on tags (v0) and another that incorporates sentiment-related keywords (v1). The use of sentiment-related keywords in v1 enhances their precision and detail.

\begin{itemize}
    \item v0, contains 12 labeling functions
\begin{lstlisting}[language=Python, caption=negatif-konflik]
@target_label(label=LABELS['negatif'])
def label_function(sample) -> bool:
  ex_tags = ['penelitian', 'penyelamatan lingkungan', 'Lembaga Swadaya Masyarakat', 'inovasi', 'trivia', 'foto']
  tags = sample['tags'].split(',')
  target_tags = ['perusahaan']

  # Implement your logic here
  if 'konflik' in tags and all(ex not in tags for ex in ex_tags):
    return True
  else:
    return False
\end{lstlisting}
\begin{lstlisting}[language=Python, caption=negatif-konflik-tambang]
@target_label(label=LABELS['negatif'])
def label_function(sample) -> bool:
  tags = sample['tags'].split(',')

  # Implement your logic here
  if 'konflik' in tags and 'tambang' in tags:
    return True
  else:
    return False
\end{lstlisting}
\begin{lstlisting}[language=Python, caption=negatif-konflik-keywords]
@target_label(label=LABELS['negatif'])
def label_function(sample) -> bool:
  ex_tags = ['penelitian', 'penyelamatan lingkungan', 'inovasi', 'trivia']
  tags = sample['tags'].split(',')
  target_tags = ['perusahaan']

  # Implement your logic here
  if 'konflik' in tags and any(target in tags for target in target_tags) and all(ex not in tags for ex in ex_tags):
    return True
  else:
    return False
\end{lstlisting}
\begin{lstlisting}[language=Python, caption=negatif-krisis-bencana]
@target_label(label=LABELS['negatif'])
def label_function(sample) -> bool:
  tags = sample['tags'].split(',')
  target_tags = ['krisis', 'bencana alam']
  ex_tags = ['penelitian', 'penyelamatan lingkungan', 'inovasi', 'trivia']

  # Implement your logic here
  for target in target_tags:
    if target in tags and all(ex not in tags for ex in ex_tags):
      return True
  return False
\end{lstlisting}
\begin{lstlisting}[language=Python, caption=positif-keywords]
@target_label(label=LABELS['positif'])
def label_function(sample) -> bool:
  ex_tags = ['konflik', 'bencana alam']
  tags = sample['tags'].split(',')
  target_tags = ['trivia', 'inovasi', 'tentang mongabay', 'penyelamatan lingkungan', 'penelitian']

  # Implement your logic here
  for target in target_tags:
    if target in tags and all(ex not in tags for ex in ex_tags):
      return True
  return False
\end{lstlisting}
\begin{lstlisting}[language=Python, caption=positif-inovasi]
@target_label(label=LABELS['positif'])
def label_function(sample) -> bool:
  tags = sample['tags'].split(',')

  # Implement your logic here
  ex_tags = ['tambang', 'bencana alam', 'korupsi']
  if 'inovasi' in tags and all(ex not in tags for ex in ex_tags):
    return True
  else:
    return False
\end{lstlisting}
\begin{lstlisting}[language=Python, caption=negatif-korupsi]
@target_label(label=LABELS['negatif'])
def label_function(sample) -> bool:
  tags = sample['tags'].split(',')

  # Implement your logic here
  if 'korupsi' in tags:
    return True
  else:
    return False
\end{lstlisting}
\begin{lstlisting}[language=Python, caption=negatif-ASN-desa]
@target_label(label=LABELS['negatif'])
def label_function(sample) -> bool:
  tags = sample['tags'].split(',')

  # Implement your logic here
  if 'Aparatur Sipil Negara' in tags and 'masyarakat desa' in tags:
    return True
  else:
    return False
\end{lstlisting}
\begin{lstlisting}[language=Python, caption=netral-keywords]
@target_label(label=LABELS['netral'])
def label_function(sample) -> bool:
  tags = sample['tags'].split(',')

  # Implement your logic here
  ex_tags = ['konflik', 'tambang', 'bencana alam', 'korupsi', 'inovasi']
  if all(ex not in tags for ex in ex_tags):
    return True
  else:
    return False
\end{lstlisting}
\begin{lstlisting}[language=Python, caption=netral-logic]
@target_label(label=LABELS['netral'])
def label_function(sample) -> bool:
  tags = sample['tags'].split(',')

  # Implement your logic here
  positif_tags = ["foto", "inovasi", "tentang mongabay", "penelitian", "penyelamatan lingkungan"]
  negatif_tags = ["bencana alam", "konflik", "korupsi", "krisis"]
  n_positif = 0
  n_negatif = 0
  for pos in positif_tags:
    if pos in tags:
      n_positif += 1
  for neg in negatif_tags:
    if neg in tags:
      n_negatif += 1

  if n_positif == n_negatif:
    return True
  else:
    return False
\end{lstlisting}
\begin{lstlisting}[language=Python, caption=positif-logic]
@target_label(label=LABELS['positif'])
def label_function(sample) -> bool:
  tags = sample['tags'].split(',')

  # Implement your logic here
  positif_tags = ["foto", "inovasi", "tentang mongabay", "penelitian", "penyelamatan lingkungan"]
  negatif_tags = ["bencana alam", "konflik", "korupsi", "krisis"]
  n_positif = 0
  n_negatif = 0
  for pos in positif_tags:
    if pos in tags:
      n_positif += 1
  for neg in negatif_tags:
    if neg in tags:
      n_negatif += 1

  if n_positif > n_negatif:
    return True
  else:
    return False
\end{lstlisting}
\begin{lstlisting}[language=Python, caption=negatif-logic]
@target_label(label=LABELS['negatif'])
def label_function(sample) -> bool:
  tags = sample['tags'].split(',')

  # Implement your logic here
  positif_tags = ["foto", "inovasi", "tentang mongabay", "penelitian", "penyelamatan lingkungan"]
  negatif_tags = ["bencana alam", "konflik", "korupsi", "krisis"]
  n_positif = 0
  n_negatif = 0
  for pos in positif_tags:
    if pos in tags:
      n_positif += 1
  for neg in negatif_tags:
    if neg in tags:
      n_negatif += 1

  if n_positif < n_negatif:
    return True
  else:
    return False
\end{lstlisting}

    \item v1, contains 16 labeling functions. In this collection, we design various rules of labeling functions, but most of them are combination of keywords with tags condition
\begin{lstlisting}[language=Python, caption=netral-abstain]
@target_label(label=LABELS['netral'])
def label_function(sample) -> bool:
  
  tags = sample['tags']

  # Implement your logic here
  if len(tags)<=1:
    return True
  else:
    return False
\end{lstlisting}
\begin{lstlisting}[language=Python, caption=netral-1tag]
# Labeling function definition (Start editing here!)
POSITIF = ['inovasi', 'penyelamatan lingkungan']
NEGATIF = ['konflik', 'krisis', 'tambang', 'korupsi', 'penyakit', 'sampah', 'hewan terancam punah']
# Assign target label based on LABELS dictionary
@target_label(label=LABELS['netral'])
def label_function(sample) -> bool:
  tags = sample['tags'].split(',')

  # Implement your logic here
  if len(tags) == 1:
    tag = tags[0]
    if tag not in POSITIF and tag not in NEGATIF:
      return True
  return False
\end{lstlisting}
\begin{lstlisting}[language=Python, caption=positif-1tag]
# Labeling function definition (Start editing here!)
POSITIF = ['inovasi', 'penyelamatan lingkungan']
NEGATIF = ['konflik', 'krisis', 'tambang', 'korupsi', 'penyakit', 'sampah', 'hewan terancam punah']
# Assign target label based on LABELS dictionary
@target_label(label=LABELS['positif'])
def label_function(sample) -> bool:
  tags = sample['tags'].split(',')

  # Implement your logic here
  if len(tags) == 1:
    tag = tags[0]
    if tag in POSITIF:
      return True
  return False
\end{lstlisting}
\begin{lstlisting}[language=Python, caption=negatif-1tag]
# Labeling function definition (Start editing here!)
POSITIF = ['inovasi', 'penyelamatan lingkungan']
NEGATIF = ['konflik', 'krisis', 'tambang', 'korupsi', 'penyakit', 'sampah', 'hewan terancam punah']
EX_KEYWORDS = ['Tahukah Anda?']
# Assign target label based on LABELS dictionary
@target_label(label=LABELS['negatif'])
def label_function(sample) -> bool:
  tags = sample['tags'].split(',')

  # Implement your logic here
  if len(tags) == 1 and all(re.search(ex, sample['title'], re.IGNORECASE)==None for ex in EX_KEYWORDS):
    tag = tags[0]
    if tag in NEGATIF:
      return True
  return False
\end{lstlisting}
\begin{lstlisting}[language=Python, caption=netral-keywords]
@target_label(label=LABELS['netral'])
def label_function(sample) -> bool:
  text = sample['title']

  # Implement your logic here
  TARGET_KEYWORDS = ['Tahukah Anda?']
  for keyword in TARGET_KEYWORDS:
    if type(keyword) == str:
      if re.search(keyword, text):
        return True
    if type(keyword) == list:
      if all(key in text for key in keyword):
        return True
  return False
\end{lstlisting}
\begin{lstlisting}[language=Python, caption=positif-inovasi]
@target_label(label=LABELS['positif'])
def label_function(sample) -> bool:
  tags = sample['tags'].split(',')
  EX_TAGS = ['kebijakan', 'konflik', 'bencana alam'] #['hewan terancam punah', 'konflik', 'pendanaan']

  # Implement your logic here
  if 'inovasi' in tags and all(ex not in tags for ex in EX_TAGS) and 'APP' not in sample['text']:
    return True
  return False
\end{lstlisting}
\begin{lstlisting}[language=Python, caption=netral-inovasi]
@target_label(label=LABELS['netral'])
def label_function(sample) -> bool:
  tags = sample['tags'].split(',')
  COMB_TAGS = ['kebijakan', 'konflik', 'bencana alam']
  EX_TAGS = [] #['hewan terancam punah', 'konflik', 'pendanaan']

  # Implement your logic here
  
  if 'inovasi' in tags and any(comb in tags for comb in COMB_TAGS):
    return True
  return False
\end{lstlisting}
\begin{lstlisting}[language=Python, caption=negatif-konflik-tags]
@target_label(label=LABELS['negatif'])
def label_function(sample) -> bool:
  tags = sample['tags'].split(',')
  PRIME_TAGS = ['konflik', 'krisis', 'korupsi']
  COMB_TAGS = ['kumpulan berita', 'tambang', 'bencana alam', 'sampah', 'lahan', 'sawit', 'penyakit']
  EX_TAGS = ['inovasi', 'penyelamatan lingkungan']
  SPECIAL_TAGS = [['tambang', 'sawit'], ['sampah', 'sawit', 'tambang', 'bencana alam']]
  SPECIAL_EX_TAGS = [['kebijakan'], ['trivia', 'Lembaga Swadaya Masyarakat']]
  EX_KEYWORDS_TITLE = ['Menyelamatkan', 'Unggulan', 'Asa ', 'Ekologi','Persatuan', 'Amankan', 'Kemandirian', 'Berdampingan', 'Penghormatan', 'Ekowisata', 'Mitigasi ', 'Si ', 'Tahukah Anda?']
  EX_KEYWORDS = ['menjaga laut', 'memuji', 'menjaga hutan', 'alhamdulillah', 'sertifikasi', 'melestarikan', 'sinergi']

  # Implement your logic here
  if len(tags) == 2:
    if any(prime in tags for prime in PRIME_TAGS):
      return True

  elif len(tags) == 3:
    if any(prime in tags for prime in PRIME_TAGS) and all(comb not in tags for comb in ['inovasi']):
      return True 

  elif any(prime in tags for prime in PRIME_TAGS) and any(key in sample['title'] for key in ['Konflik', 'Ironi', 'Polemik', 'Penyumbang Masalah', 'Belum Aman']):
      return True

  n = 0
  for tag in tags:
    if tag in PRIME_TAGS:
      n += 1

  if all(ex not in tags for ex in EX_TAGS) and all(ex not in sample['title'] for ex in EX_KEYWORDS_TITLE):
    for s_tag, s_ex in zip(SPECIAL_TAGS, SPECIAL_EX_TAGS):
      if any(prime in tags for prime in PRIME_TAGS) and all(special not in tags for special in s_tag) and all(ex in tags for ex in s_ex):
        return False
      elif any(prime in tags for prime in PRIME_TAGS) and any(tag in tags for tag in s_tag) and all(ex in tags for ex in s_ex):
        if any(ex_kw in sample['text'].lower() for ex_kw in EX_KEYWORDS):
          return False
    
    if n>=2:
      return True
    elif n ==1 and any(comb in tags for comb in COMB_TAGS):
      return True
    elif n==1 and 'tambang' in tags and 'penyelamatan lingkungan' in tags:
      return True

  return False
\end{lstlisting}
\begin{lstlisting}[language=Python, caption=negatif-korupsi]
@target_label(label=LABELS['negatif'])
def label_function(sample) -> bool:
  tags = sample['tags'].split(',')
  PRIME_TAGS = ['korupsi']
  EX_TAGS = ['inovasi', 'tentang mongabay']
  EX_KEYWORDS = ['mengurangi korupsi', 'Tahukah anda?']

  if any(prime in tags for prime in PRIME_TAGS) and all(ex not in tags for ex in EX_TAGS) and all(ex not in sample['title']+' | '+sample['text'] for ex in EX_KEYWORDS):
    return True
  return False
\end{lstlisting}
\begin{lstlisting}[language=Python, caption=negatif-konflik]
@target_label(label=LABELS['negatif'])
def label_function(sample) -> bool:
  tags = sample['tags'].split(',')
  PRIME_TAGS = ['konflik']
  COMBS = [['bencana alam', 'lahan'], ['bencana alam', 'masyarakat desa'],['bencana alam', 'sampah'],['hewan terancam punah'],['perdagangan', 'hewan terancam punah'],['pendanaan'],['Aparatur Sipil Negara','masyarakat desa'],['perusahaan', 'masyarakat desa'],['Aparatur Sipil Negara','perusahaan'],['Aparatur Sipil Negara','tambang']]
  EX_TAGS = ['inovasi', 'tentang mongabay'] #'penyelamatan lingkungan']
  EX_KEYWORDS_TITLE = ['Menyelamatkan', 'Ekologi', ' Asa ', 'Ajak', 'Semangat', 'Kisah', 'Pentingnya', 'Pelestarian', 'Defender', 'Menghentikan Tambang', 'Persatuan', 'Amankan', 'Kemandirian', 'Berdampingan', 'Penghormatan', 'Ekowisata', 'Mitigasi ', 'Si ', 'Tahukah Anda?']
  EX_KEYWORDS = [] #['menjaga laut', 'memuji', 'menjaga hutan', 'alhamdulillah', 'sertifikasi', 'melestarikan', 'sinergi']

  # Implement your logic here
  for comb in COMBS:
    if any(prime in tags for prime in PRIME_TAGS) and all(c in tags for c in comb) and all(ex not in tags for ex in EX_TAGS) and all(ex not in sample['title'] for ex in EX_KEYWORDS_TITLE) and all(ex not in sample['text'] for ex in EX_KEYWORDS):
      return True
    elif any(prime in tags for prime in PRIME_TAGS) and 'Konflik' in sample['title']:
      return True
  return False
\end{lstlisting}
\begin{lstlisting}[language=Python, caption=netral-konflik]
@target_label(label=LABELS['netral'])
def label_function(sample) -> bool:
  tags = sample['tags'].split(',')
  PRIME_TAGS = ['konflik', 'krisis']
  COMBS = [['kebijakan','politik'],['kebijakan','perdagangan'],['kebijakan', 'Lembaga Swadaya Masyarakat'], ['penyelamatan lingkungan'],['kebijakan','budidaya']]
  EX_COMBS = [['bencana alam', 'lahan'], ['bencana alam', 'masyarakat desa'],['bencana alam', 'sampah'],['hewan terancam punah'],['perdagangan', 'hewan terancam punah'],['pendanaan'],['Aparatur Sipil Negara','masyarakat desa'],['perusahaan', 'masyarakat desa'],['Aparatur Sipil Negara','perusahaan'],['Aparatur Sipil Negara','tambang'],['Aparatur Sipil Negara','Lembaga Swadaya Masyarakat']]
  EX_KEYWORDS = ['Belum Aman', 'Penyumbang Masalah', 'Polemik', 'Ironi', 'Konflik', 'Intimidasi', 'Perompakan', 'Lestari', 'Unggulan', 'Ajak', 'Semangat', 'Kisah', 'Pentingnya', 'Pelestarian', 'Defender', 'Menghentikan Tambang', 'Persatuan', 'Amankan', 'Kemandirian', 'Berdampingan', 'Penghormatan']
  EX_TAGS = ['inovasi', 'korupsi', 'tambang', 'krisis', 'hewan terancam punah'] #['hewan terancam punah', 'konflik', 'pendanaan']

  # Implement your logic here
  if all(ex not in tags for ex in EX_TAGS) and all(ex not in sample['title'] for ex in EX_KEYWORDS):
    if all(all(ex_c not in tags for ex_c in ex_comb) and any(prime in tags for prime in PRIME_TAGS) for ex_comb in EX_COMBS):
      for comb in COMBS:
        if any(prime in tags for prime in PRIME_TAGS) and all(c in tags for c in comb):
          return True
  return False
\end{lstlisting}
\begin{lstlisting}[language=Python, caption=positif-konflik]
@target_label(label=LABELS['positif'])
def label_function(sample) -> bool:
  tags = sample['tags'].split(',')
  COMBS = [] #['kumpulan berita', 'tambang', 'bencana alam', 'sampah', 'lahan', 'sawit']
  PRIME_TAGS = ['konflik', 'krisis']
  KEYWORDS = ['Lestari', 'Unggulan', 'Ajak', 'Semangat', 'Kisah', 'Pentingnya', 'Pelestarian', 'Defender', 'Menghentikan Tambang', 'Persatuan', 'Amankan', 'Kemandirian', 'Berdampingan', 'Penghormatan']
  # Implement your logic here
  if any(prime in tags for prime in PRIME_TAGS) and any(key in sample['title'] for key in KEYWORDS):
    return True
  return False
\end{lstlisting}
\begin{lstlisting}[language=Python, caption=negatif-krisis]
@target_label(label=LABELS['negatif'])
def label_function(sample) -> bool:
  tags = sample['tags'].split(',')
  PRIME_TAGS = ['krisis']
  EX_TAGS = ['konflik', 'inovasi', 'tentang mongabay']
  EX_KEYWORDS = []

  if any(prime in tags for prime in PRIME_TAGS) and all(ex not in tags for ex in EX_TAGS) and all(ex not in sample['title']+' | '+sample['text'] for ex in EX_KEYWORDS):
    if any(tag in tags for tag in ['politik', 'pendanaan', 'tambang', 'sawit', 'lahan']):
      return True
  return False
\end{lstlisting}
\begin{lstlisting}[language=Python, caption=positif-tags]
@target_label(label=LABELS['positif'])
def label_function(sample) -> bool:
  tags = sample['tags'].split(',')
  EX_TAGS = ['konflik', 'krisis', 'korupsi', 'sampah', 'inovasi']
  POS_TAGS = [['penyelamatan lingkungan', 'Lembaga Swadaya Masyarakat'], ['penyelamatan lingkungan', 'pertanian'],['penyelamatan lingkungan', 'hewan terancam punah']]
  POS_EX_TAGS = [['hewan terancam punah', 'kebijakan', 'perdagangan', 'pendanaan', 'sawit', 'tambang'], ['hewan terancam punah', 'kebijakan', 'perdagangan', 'pendanaan', 'tambang'], ['kebijakan', 'perdagangan', 'pendanaan', 'tambang']]
  NEG_TAGS = [['perdagangan', 'hewan terancam punah'], ['tambang']]
  NETRAL_TAGS = ['hewan terancam punah', 'kebijakan', 'perdagangan', 'pendanaan']
  
  if all(tag not in EX_TAGS for tag in tags):
    if any(all(pos in tags for pos in pos_tags) and all(ex not in tags for ex in pos_ex_tags) for (pos_tags, pos_ex_tags) in zip(POS_TAGS, POS_EX_TAGS)):
      return True
    elif any(all(neg in tags for neg in neg_tags) for neg_tags in NEG_TAGS):
      return False
    else:
      return False
  return False
\end{lstlisting}
\begin{lstlisting}[language=Python, caption=negatif-tags]
@target_label(label=LABELS['negatif'])
def label_function(sample) -> bool:
  tags = sample['tags'].split(',')
  EX_TAGS = ['konflik', 'krisis', 'korupsi', 'sampah', 'inovasi']
  POS_TAGS = [['penyelamatan lingkungan', 'Lembaga Swadaya Masyarakat'], ['penyelamatan lingkungan', 'pertanian'],['penyelamatan lingkungan', 'hewan terancam punah']]
  POS_EX_TAGS = [['hewan terancam punah', 'kebijakan', 'perdagangan', 'pendanaan', 'sawit', 'tambang'], ['hewan terancam punah', 'kebijakan', 'perdagangan', 'pendanaan', 'tambang'], ['kebijakan', 'perdagangan', 'pendanaan', 'tambang']]
  NEG_TAGS = [['perdagangan', 'hewan terancam punah'], ['tambang']]
  NETRAL_TAGS = ['hewan terancam punah', 'kebijakan', 'perdagangan', 'pendanaan']
  
  if all(tag not in EX_TAGS for tag in tags):
    if any(all(pos in tags for pos in pos_tags) and all(ex not in tags for ex in pos_ex_tags) for (pos_tags, pos_ex_tags) in zip(POS_TAGS, POS_EX_TAGS)):
      return False
    elif any(all(neg in tags for neg in neg_tags) for neg_tags in NEG_TAGS):
      return True
    else:
      return False
  return False
\end{lstlisting}
\begin{lstlisting}[language=Python, caption=netral-tags]
@target_label(label=LABELS['netral'])
def label_function(sample) -> bool:
  tags = sample['tags'].split(',')
  EX_TAGS = ['konflik', 'krisis', 'korupsi', 'sampah', 'inovasi']
  POS_TAGS = [['penyelamatan lingkungan', 'Lembaga Swadaya Masyarakat'], ['penyelamatan lingkungan', 'pertanian'],['penyelamatan lingkungan', 'hewan terancam punah']]
  POS_EX_TAGS = [['hewan terancam punah', 'kebijakan', 'perdagangan', 'pendanaan', 'sawit', 'tambang'], ['hewan terancam punah', 'kebijakan', 'perdagangan', 'pendanaan', 'tambang'], ['kebijakan', 'perdagangan', 'pendanaan', 'tambang']]
  NEG_TAGS = [['perdagangan', 'hewan terancam punah'], ['tambang']]
  NETRAL_TAGS = ['hewan terancam punah', 'kebijakan', 'perdagangan', 'pendanaan']
  
  if all(tag not in EX_TAGS for tag in tags):
    if any(all(pos in tags for pos in pos_tags) and all(ex not in tags for ex in pos_ex_tags) for (pos_tags, pos_ex_tags) in zip(POS_TAGS, POS_EX_TAGS)):
      return False
    elif any(all(neg in tags for neg in neg_tags) for neg_tags in NEG_TAGS):
      return False
    else:
      return True
  return False
\end{lstlisting}
\end{itemize}

\twocolumn

\subsection{Hyperparameter}

\begin{table}[h]
\begin{minipage}[h]{\linewidth}
\centering
\begin{tabular}{|c|c|}
\hline
\textbf{Hyper-Params} & \textbf{Value} \\ \hline
optimizer & Adam \\
l2 regularization & 0.01 \\
epoch & 2 \\
batch size & 5000 \\
lr & 1e-2 \\\hline
\end{tabular}
\caption{Label model training Hyperparams}
\label{tab:label_model_hp}
\end{minipage}
\end{table}

\begin{table}[h]
\begin{minipage}[h]{\linewidth}
\centering
\begin{tabular}{|c|c|c|}
\hline
\textbf{Hyper-Params} & \textbf{Tags} & \textbf{Sentiment}\\ \hline
epoch & 20 & 20\\
train batch size & 32 & 32\\
eval batch size & 1 & 1\\
optimizer & Adam & Adam\\
adam beta1 & 0.9 & 0.9\\
adam beta 2 & 0.95 & 0.95\\
adam epsilon & 1e-8 & 1e-8\\
lr & 1e-6 & 1e-4 \\
warmup ratio & 0.03 & 0.03 \\
lr scheduler type & cosine,linear & cosine,linear \\ \hline
\end{tabular}
\caption{Model finetuning Hyperparams}
\label{tab:finetuning_hp}
\end{minipage}
\end{table}

\subsubsection{F1-Score per label}
\label{sec:f1_score_label}
\begin{table}[h]
\begin{tabular}{|c||l|l|l|l|l|l|}
  \hline
  \multirow{2}{*}{Label} 
      & \multicolumn{2}{c|}{indobert} 
          & \multicolumn{2}{c|}{mbert}
            & \multicolumn{2}{c|}{bert}\\             \cline{2-7}
  & CM & MV & CM & MV & CM & MV \\  \hline
  positive & 44.1 & 37.5 & 24.8 & 0 & 7 & 0 \\      \hline
  neutral & 46.3 & 48.2 & 37.4 & 44.3 & 38.5 & 42.5 \\      \hline
  negative & 76.7 & 76.7 & 70.3 & 70.4 & 61.9 & 57.1\\      \hline
\end{tabular}
\caption{F1-score per label for sentiment classification task}
\end{table}
\begin{table}
\begin{tabular}{|c||l|l|l|l|l|l|}
  \hline
  \multirow{2}{*}{Tag} 
      & \multicolumn{2}{c|}{indobert} 
          & \multicolumn{2}{c|}{mbert}
            & \multicolumn{2}{c|}{bert}\\             \cline{2-7}
  & CM & MV & CM & MV & CM & MV \\  \hline
  ASN & 84.5 & 88.1 & 83.3 & 85.6 & 56.3 & 54.5 \\      \hline
  LSM & 69.3 & 72.6 & 71.9 & 74.4 & 41 & 47.5 \\      \hline
  bencana & 82.4 & 86.1 & 83.8 & 82.7 & 36.5 & 42.7\\      \hline
  budidaya & 50.5 & 51.7 & 47.3 & 48.6 & 36.7 & 43.6\\      \hline
  desa & 42.9 & 43.6 & 42.1 & 43.5 & 42.1 & 44.6\\      \hline
  energi & 73.5 & 70.8 & 75.6 & 65.1 & 59.1 & 65\\      \hline
  foto & 55.8 & 55.8 & 50 & 59.1 & 64.9 & 65\\      \hline
  iklim/cuaca & 87.6 & 90.1 & 84.1 & 86 & 59.5 & 64.9\\      \hline
  inovasi & 48.1 & 52.3 & 21.7 & 41.5 & 27.1 & 25.3\\      \hline
  kebijakan & 46.9 & 48.5 & 44.9 & 49.1 & 42 & 48.4\\      \hline
  konflik & 77.3 & 80.2 & 73.4 & 79.5 & 62.9 & 63\\      \hline
  korupsi & 32.3 & 37.5 & 36.4 & 36.4 & 0 & 27.6\\      \hline
  krisis & 78 & 76.2 & 74.4 & 76.2 & 40.9 & 50\\      \hline
  news-bulk & 0 & 0 & 0 & 0 & 0 & 0\\      \hline
  lahan & 82.7 & 85.6 & 77.8 & 87.3 & 53.2 & 62.3\\      \hline
  mangrove & 94.7 & 94.7 & 88.9 & 85.7 & 94.7 & 90\\      \hline
  nelayan & 0 & 0 & 0 & 0 & 0 & 0\\      \hline
  mongabay & 88.4 & 89.7 & 89.7 & 89.7 & 51.5 & 73.4\\      \hline
  dana & 69.2 & 70.4 & 65.8 & 67.5 & 32.1 & 50.4\\      \hline
  penelitian & 72 & 67.7 & 68.7 & 69.8 & 58.8 & 57.4\\      \hline
  penyakit & 61.4 & 62.6 & 63.3 & 60.8 & 23.2 & 41\\      \hline
  go-green & 41.4 & 43.6 & 40.5 & 46.9 & 33.7 & 43.6\\      \hline
  perdagangan & 75.6 & 73.9 & 71.4 & 74.5 & 45.9 & 57.9\\      \hline
  pertanian & 85.2 & 84.6 & 79.3 & 81.3 & 49.5 & 62.9\\      \hline
  perusahaan & 78.3 & 83.3 & 78.2 & 83.4 & 58.4 & 68.8\\      \hline
  politik & 70.5 & 71.5 & 70 & 70.4 & 61.8 & 63.2\\      \hline
  end-animal & 68.2 & 65.6 & 43.5 & 49.3 & 38.4 & 52.3\\      \hline
  sampah & 76.9 & 79.5 & 67.5 & 78.5 & 20.7 & 41.4\\      \hline
  sawit & 89.8 & 93.2 & 88.4 & 94 & 75.2 & 85.7\\      \hline
  tambang & 89.3 & 92.8 & 89.7 & 93.5 & 55.7 & 72.1\\      \hline
  trivia & 64.3 & 60.4 & 56.3 & 53.2 & 56 & 46.2\\      \hline
\end{tabular}
\caption{F1-score per tag for tags classification task}
\end{table}
\onecolumn

\subsection{Tags Explanation}
\label{sec:tags_explanation}
\begin{table*}[!ht]
\begin{tabular}{|l|l|}
\hline \textbf{Tags} & \textbf{Explanation} \\  \hline
ASN & Civil Service Agency that took part in the event of the news \\  \hline
LSM & Non-Governmental Organizations (NGO) that took action in the \\ & event of the news (commonly in go-green/conservation action)\\  \hline
bencana & disaster that was discussed in the news\\  \hline
budidaya & plant cultivation/fishery that was discussed in the news\\  \hline
desa & villager/village life that becomes the subject of the news\\  \hline
energi & energy issue/solution that was discussed in the news\\  \hline
foto & short news that discusses the photo, commonly the uniqueness of ethnic \\ & animals/culture/moment that was captured in the photo\\ \hline 
iklim/cuaca & weather/climate issues that discussed in the news\\ \hline
inovasi & innovation of an action/product/rule that was discussed in the news \\ & (commonly comes from citizens) \\ \hline
kebijakan & policy related to conservation or environment that commonly comes up \\ & from government\\ \hline
konflik & conflict between two or more parties that were included/narrated in the news\\ \hline
korupsi & corruption/bribery case that was discussed in the news\\ & (commonly related to manipulation of conservation funding) \\ \hline
krisis & crisis (environment/weather/hunger/war) that was discussed in the news \\ & (commonly was narrated from a third-party point of view, sometimes \\ & it was the indirect impact of conflict/disaster\\ \hline
kumpulan berita & bulk of news articles from various topics that were delivered as \\ & a long-list article\\ \hline
lahan & regulatory of parts of land that were discussed in the news (commonly \\ & narrated as a conflict between regulatory/company and citizens/NGO)\\ \hline
mangrove & mangrove topics/action/issue that was discussed in the news\\ \hline
nelayan & fisherman life/conflict/related-policy that was discussed in the news\\ \hline
mongabay & news that raised redaction (Mongabay) action/event/concern related \\ & to conservation\\ \hline
dana & funding (in relation to conservation) that was discussed in the news\\ \hline
penelitian & research/exploration in nature field (solution proposal/announce\\ & newest species/point out risk analysis) that was discussed in the news.\\ \hline
penyakit & disease that become impact/concern/impact in the news\\ \hline
go-green & conservation/go-green/endangered-animal-saving related action/campaign,\\ & commonly proposed by NGO/citizens\\ \hline
perdagangan & trading activities related to conservation that was pointed out in the news\\ \hline
pertanian & agriculture issue/solution/concern that was discussed in the news\\ \hline
perusahaan & company/corporation that was involved in conservation-related \\ & regulation/activities\\ \hline
politik & politics/government activities/regulation/candidacy/scandal\\ & that were mentioned in the news\\ \hline
endangered animal & endangered animal that was concerned in the news\\ \hline
sampah & waste issue/solution/action that was discussed in the news\\ \hline
sawit & any event/regulation/conflict/solution related to palm oil\\ & (in terms of one of fuel resources)\\ \hline
tambang & any event/regulation/conflict/solution related to mining activities\\ \hline
trivia & trivia/non-serious event/topic that was raised in the news \\ & (commonly related to fun-fact, biography, culture activities, etc)\\ \hline
\end{tabular}
\end{table*}



\end{document}